\documentclass[a4paper]{article}

\usepackage{INTERSPEECH2021}

\usepackage[color=white]{todonotes}  
\usepackage{adjustbox}
\usepackage{multirow}
\usepackage{tikz}
\usepackage{pgfplots}
\usepackage{enumitem}
\usepackage{hyperref}
\usepackage{verbatim}

\newcommand{\mtit}[1]{\mathit{#1}}

\title{Information Retrieval for ZeroSpeech 2021:\\The Submission by University of Wroclaw}
\name{Jan Chorowski$^{1,2}$, Grzegorz Ciesielski$^1$, Jarosław Dzikowski$^1$, Adrian Łańcucki$^3$, Ricard Marxer$^4$, Mateusz Opala$^1$, Piotr Pusz$^1$, Paweł Rychlikowski$^1$ and Michał Stypułkowski$^1$}
\address{
  $^1$University of Wroclaw, Poland\\
  $^2$NavAlgo, France\\
  $^3$NVIDIA\\
  $^4$Université de Toulon, Aix Marseille Univ, CNRS, LIS, France}
\email{jan.chorowski@cs.uni.wroc.pl, pawel.rychlikowski@cs.uni.wroc.pl}

\begin{document}
\maketitle
\begin{abstract}
  We present a number of low-resource approaches to the tasks of the
  Zero Resource Speech Challenge 2021. We build on the unsupervised representations of speech proposed by the organizers as a baseline,
  derived from CPC and clustered with the k-means algorithm.
  We demonstrate that simple methods of refining those representations can narrow the gap,
  or even improve upon the solutions which use a high computational budget.
  The results lead to the conclusion that the CPC-derived representations
  are still too noisy for training language models, but stable enough for simpler forms of pattern matching and retrieval.
\end{abstract}

\section{Introduction}
The Zero Resource Speech Challenge series (ZeroSpeech)~\cite{versteegh2015zero,nguyen_zero_2020} is an initiative with the ultimate goal of building from scratch a system that learns an end-to-end spoken dialogue system for an unknown language, using only sensory information mainly in the form of recordings, and does not use any linguistic resources or knowledge.


The high-level objective of the competition is to learn various qualities of a natural language at different levels of granularity
directly from raw audio without any supervision.
ZeroSpeech 2021 evaluates speech understanding using the following tasks and datasets:
\begin{enumerate}[noitemsep,topsep=0pt] 
     \item Phonetic ABX (Libri-Light dataset~\cite{kahn2020libri}), where the system has to judge whether two phonemes are identical
     \item Lexical (sWUGGY dataset) - classifying whether a spoken utterance is a real or misspelled word
     \item Semantic (sSIMI dataset) - assessing semantic similarity between two spoken words
     \item Syntactic (sBLIMP dataset) - assigning a \textit{grammatical score} to an utterance in such a way that erroneous sentences have lower scores than correct ones
 \end{enumerate}
For all tasks it is assumed that spoken corpora (either LibriSpeech~\cite{7178964}, or Libri-Light~\cite{kahn2020libri}) are the only sources of language knowledge.



%
%

The organizers have provided a baseline solution~\cite{nguyen_zero_2020}, which we adapt and modify in our submission. 
Raw audio is 
fed to a Contrastive Predicting Coding (CPC) model~\cite{oord_representation_2018}. By trying to predict future representations, the CPC model learns useful intermediate representations of speech. These come in the form of an embedding vector emitted every 10 ms. Collected from a large training corpus, the embeddings are then clustered with the k-means algorithm into 50 pseudo-phones. With this pipeline, any unseen audio can be transformed into a stream of pseudo-phones, on which language models may be trained for downstream tasks.

In this paper we present our submission which tries to address all four tasks. We extend the baseline solution in several directions: we refine the intermediate representations, extracted with CPC, to directly improve the ABX scores. We show that such representations can be used to perform simple fuzzy look-ups in a large dataset, and even extract some common patterns that serve as pseudo-words.
Our approach to the semantic word similarity task is also based on pseudo-words. Instead of pooling the hidden states of the language model, we opt for a direct discovery of pseudo-words in the corpus. These can be embedded with a word2vec-like approach~\cite{mikolov2013efficient} to form a semantic vector for the entire word. 
Lastly, for the syntax modeling task we use a simple LSTM model similar to the baseline one.

We provide complete source code of our submission at \url{https://github.com/chorowski-lab/zs2021}.


\section{Phonetic Task: Libri-Light ABX}
In the ABX task two speech categories A and B are determined by two recordings (e.g., ``bit'' vs ``bet''), and a third recording X has to be correctly assigned to either one of them.

The baseline representations for the ABX task are CPC-derived embedding vectors.
In order to improve upon those representations, we introduce two approaches described in the following section.

\subsection{Improvements to CPC Representations}

\noindent\textbf{Factoring Out Speaker Identities }
The embeddings produced by CPC contain information about both the phonetic content and speaker identity. In case of ABX, which is a phoneme recognition metric, the latter is irrelevant. We therefore project the embeddings of the baseline model (CPC-big~\cite{nguyen_zero_2020}) into the nullspace of a linear speaker classification model to render the embeddings less speaker-sensitive. We perform speaker classification on baseline CPC embeddings with a projection factorized into matrices $A$ and $B$, where $A \in \mathbb{R}^{D_\mtit{inb} \times D_\mtit{emb}}$, $B \in \mathbb{R}^{D_\mtit{spk} \times D_\mtit{inb}}$, $D_\mtit{emb}$ is the dimensionality of embeddings and $D_\mtit{inb}$ is the linear bottleneck dimensionality. In order to compute ABX, we multiply the CPC-derived embeddings by $A'\in \mathbb{R}^{(D_\mtit{emb} - D_\mtit{inb}) \times D_\mtit{emb}}$, the nullspace matrix of $A$.\\

\noindent\textbf{Averaging with Centroids }
%
Higher-level tasks of the competition rely on pseudo-phones found by clustering these vectors with k-means. Doing so proves useful, so we incorporate some of the outcomes of the clustering back into the dense CPC-derived vectors.

Specifically, we take a weighted average of every dense CPC-derived embedding $e$ in the embedding space with its cluster centroid $c_e$:
\begin{equation}
    \hat{e} = \alpha \ c_e + (1-\alpha)\ e.
\end{equation}
This averaging moves every dense embedding towards its assigned centroid proportionally to the distance from it. This aims to include information about the global characteristics of the embedding space coming from clustering without substantial loss of local information, and might be regarded as a simple form of denoising.
It does not change the assignment to the closest centroid.

\subsection{ABX Experiments}
\label{sec:abx}
We evaluate both aforementioned improvements on the ZeroSpeech ABX task.
To begin with, we extract 512-dimensional embeddings from the second layer of the CPC's autoregressive module. We run each classification experiment for 10 epochs on the \emph{train-clean-100} subset of LibriSpeech.

ZeroSpeech 2021 dev/test sets are subsets of LibriSpeech dev/test sets, respectively. However, for best results (and for replication of the baseline) we had to first compute the embeddings on the full LibriSpeech test set, to allow the model to keep latent state between consecutive utterances. After all embeddigs were computed, we have kept only the ones needed for ZeroSpeech ABX evaluations. 


We investigate the variation in ABX scores with respect to the dimensionality of the resulting nullspace, by testing with different bottleneck sizes $D_{\mtit{inb}}$ of the speaker classifier.
%
We achieve the best ABX result when the nullspace size is 448. 

Next, we evaluate the influence of averaging with centroids on the ABX scores (Table \ref{tab:centerpush_abx}).
It also noticeably improves ABX results, and we achieve the highest error reduction when it is combined with a 448-dimensional nullspace projection.
We have experimented with both Euclidean and cosine distance metrics when performing k-means and later choosing the closest centroid. Both yield similar results, and we select the centroid according to the cosine distance in subsequent experiments.

\begin{table}[tb]
\centering
  \caption{ABX error rates (\%, cosine distance) for multiple sizes of nullspaces of speaker classification models. 
  The nullspace dimension complements the bottleneck dimension used to train the speaker recognizer.}
  \label{tab:nullspaces_abx}
\begingroup\raggedleft
\noindent\adjustbox{max width=\columnwidth}{

\begin{tabular}{ l c c c c c c c c}
\toprule
         & &                                  & \multicolumn{6}{c}{\bf Nullspace dimensionality}\\
\cmidrule{4-9}
\multicolumn{3}{l}{\bf Evaluation} & \bf None & \bf 464 & \bf 448 & \bf 416 & \bf 320 & \bf 256 \\
\midrule
 dev   & clean & within & 3.38 & 3.28 & {\bf 3.25} & 3.29 & 3.26 & 3.31 \\
       & clean & across & 4.17 & 3.98 & {\bf 3.94} & 3.92 & 3.98 & 3.99 \\
       & other & within & 4.81 & 4.63 & {\bf 4.60} & 4.61 & 4.62 & 4.67 \\
       & other & across & 7.53 & 7.34 & {\bf 7.24} & 7.24 & 7.21 & 7.26 \\
 \bottomrule
\end{tabular}%
}
\endgroup
\vspace{0mm}
\end{table}

\begin{table}[tb]
  \centering
  \caption{Phoneme and speaker classification accuracies (\%) of models after applying factorized projection heads (top) and nullspace matrices of the aforementioned speaker classification models (bottom) on the baseline embeddings (c.f. Table~\ref{tab:nullspaces_abx}).}
  \label{tab:nullspaces_acc}%
  \noindent\adjustbox{max width=\columnwidth}{
  \begin{tabular}{l*{8}{c}}
    \toprule
              & \multicolumn{6}{c}{\bf Proj. head bottleneck / Nullspace dimensionality}\\
    \cmidrule{2-7}
    \bf Setup                            & \bf None / 512 & \bf 48 / 464 & \bf 64 / 448 & \bf 96 / 416 & \bf 192 / 320 & \bf 256 / 256 \\
	\midrule
	Speakers (fact. proj. head)          & 84.85 & 91.17 & 91.32 & 91.59 & 92.02 & 92.02 \\ \midrule
    Speakers + null space                & 84.85 & 29.42 & 27.91 & 24.95 & 19.50 & 15.56 \\
	Phonemes + null space                & 78.61 & 78.46 & 78.36 & 78.18 & 77.46 & 76.86 \\ 
	\bottomrule
  \end{tabular}
    }
\end{table}

\begin{table}[tb]
\caption{ABX error rates (\%, cosine distance) for weighted averaging of CPC embeddings with centroids. The bottom half shows results combined with the best 448-dimensional nullspace setup. The nullspace dimension is equal to the difference of dimensions between the embeddings and the bottleneck used to train the speaker classifier. Phoneme classification results in Table \ref{tab:centerpush_linsep}}
\label{tab:centerpush_abx}
\begingroup\raggedleft
\noindent\adjustbox{max width=\columnwidth}{
\begin{tabular}{ l c c c c c c c c } 
\toprule
& & & \multicolumn{6}{c}{\bf Centroid weight}\\
\cmidrule{4-9}
\multicolumn{2}{l}{\bf Evaluation} & & \bf None & {\bf 0.2} & {\bf 0.3} & {\bf 0.4} & {\bf 0.5} & {\bf 0.6} \\ 
\toprule
dev                       & clean & within & 3.43 & 3.23 & 3.16 & \textbf{3.09} & 3.07 & 3.22 \\ 
                          & clean & across & 4.20 & 3.96 & 3.84 & \textbf{3.76} & 3.77 & 3.97 \\ 
                          & other & within & 4.84 & 4.64 & 4.62 & \textbf{4.61} & 4.75 & 5.19 \\ 
                          & other & across & 7.63 & 7.38 & 7.28 & \textbf{7.32} & 7.53 & 7.93 \\ 
\midrule
dev                       & clean & within & 3.25 & 3.03 & 2.97 & \textbf{2.94} & 2.93 & 3.10\\ 
 + nullspace              & clean & across & 3.94 & 3.66 & 3.60 & \textbf{3.58} & 3.57 & 3.75\\ 
                          & other & within & 4.60 & 4.49 & 4.47 & \textbf{4.47} & 4.66 & 4.98\\ 
                          & other & across & 7.24 & 7.05 & 6.94 & \textbf{7.02} & 7.21 & 7.67\\ 
\bottomrule
\end{tabular}%
}
\endgroup
\vspace{0mm}
\end{table}

Lastly, we evaluate the influence of both methods on supervised speaker and phoneme classification accuracies.
In the nullspace approach, both of them are low (Table~\ref{tab:nullspaces_acc}), and increase with the size of the nullspace ($D_\mtit{emb} - D_\mtit{inb}$).
This indicates that after we attempt to make the representations speaker-insensitive, there is still some stray of speaker-related information present in the remaining dimensions.
As seen in Table \ref{tab:centerpush_linsep}, averaging with centroids also reduces phoneme classification accuracy, proportionally to how much the embeddings are altered, both with and without the nullspace projection.

Thus, both tested methods improve ABX, and degrade phoneme separation results. This can be because difficulties of those tasks and power of downstream models used differs - we discard less important parts of the information, which improves ABX results as the task is simple and there are no trained parameters, just representation distances (in which case removing less important parts of information helps) but degrades phoneme separation performance (as we still discard some information which classifier with trainable weights could use). In both cases the relative gain in ABX is bigger than the relative loss in phoneme separation performance.

\begin{table}[htb]
  \centering
  \caption{Phoneme classification accuracies (\%) for averaging with centroids, both without the nullspace and after projection to the nullspace. ABX results in Table \ref{tab:centerpush_abx}} 
  \label{tab:centerpush_linsep}
  \noindent\adjustbox{max width=\columnwidth}{
  \begin{tabular}{lcccccc}
  \toprule
    & \multicolumn{6}{c}{\bf Centroid weight}\\
    \cmidrule{2-7}
    Euclidean k-means          & \bf None & {\bf 0.2} & {\bf 0.3} & {\bf 0.4} & {\bf 0.5} & {\bf 0.6} \\ 
  \midrule
	Phonemes                   & \textbf{78.0} & 77.6 & 77.3 & 76.9 & 76.3 & 75.6 \\ 
	Phonemes + 448-d nullspace & 77.7 & 77.4 & 77.0 & 76.6 & 76.0 & 75.4 \\ 
	\bottomrule
  \end{tabular}
   }
\vspace{0mm}
\end{table}

\section{Quantization for Higher-level Tasks}
The baseline methods for the remaining tasks of higher linguistic levels require quantized inputs, that act as discrete input tokens for language models.
This is achieved by clustering CPC-derived vectors with k-means, and mapping every dense vector to its centroid.
To achieve the best results on lexical and syntactic tasks (sWUGGY and sBLIMP datasets), we use the CPC-nullspace embeddings instead of the raw CPC embeddings.
In contrast to feature extraction for the ABX task, now the LSTM context in CPC is not kept between the files, as the datasets for specific tasks are not related to one another.
Additionally, when computing the distances, we normalize $L_2$-norm lengths of vectors and in effect switch from the Euclidean metric to the cosine metric for quantization.

For the semantic task (sSIMI dataset), we use the baseline quantizations produced with the raw CPC embeddings, and k-means clustering with the Euclidean metric.

\section{Lexical Task: sWUGGY}
The goal of the task is to distinguish a real word from a similar pseudo-word in a given pair.
Pseudo-words were generated with Wuggy~\cite{keuleers2010wuggy}, and adhere to orthographic and phonological patterns of the English language. Such pairs make up the sWUGGY dataset, which has two parts: one with real words which appear in the LibriSpeech training set (\emph{base}), and another one in which they do not (\emph{OOV}).

The baseline solution takes pseudo-phones as input, and judges the likelihood of a word with a language model, following \cite{le-godais-etal-2017-comparing}.
For the \emph{base} pairs, our method performs a dictionary lookup of the quantized representations trying to spot the words in the entire LibriSpeech training corpus.

For the \emph{OOV} pairs, we were trying to use a simple LSTM language model and to combine it with dictionary lookup. But since guessing whether a word is in vocabulary is somehow problematic, and LSTM yielded similar results to DTW, we decided to treat all words in the same way.

\subsection{Dictionary Lookup}
We build a corpus by pre-processing all LibriSpeech training set utterances to strings of pseudo-phones.
For every query word, our goal is to find the closest matching subsequence in the corpus.
The lookup is based on dynamic time warping (DTW)~\cite{1163055}, 
and uses subsequence DTW which matches a given sequence to a contiguous subsequence of another, such that the matching cost is minimal across all subsequences.
This can be done without any increase in complexity, and is easy to parallelize.

Query words and pseudo-phone representations of training utterances are strings of discrete centroid numbers. A simple similarity matrix between the elements of two sequences $x, y$ would be a binary one. We take advantage of having Euclidean coordinates of the centroids, and compare two pseudo-phones by the similarity of their centroid vectors. Thus, every similarity matrix has real-valued entries, and we perform soft matching of sequences.

We estimate pseudo-log probability of a query word as the negative quotient of the minimum DTW cost to the mean DTW cost of this word.
We normalize with the mean DTW cost because longer words tend to have higher costs, which would result in a bias towards shorter words.


\subsection{Experiments for sWUGGY}
\begin{table}[htb]
  \centering
  \caption{Scores for DTW lookup on different quantizations, and with linear and optimal distance matrices. The results were computed for the base dev set of the lexical task (no OOV subset). We used the train-full-960 subset of the LibriSpeech as dictionary.}
  \label{tab:swuggy_dtw}
  \begin{tabular}{llcc}
    \toprule
	 &  & \multicolumn{2}{c}{\textbf{Classification accuracy}} \\
	\textbf{Quantization} & \textbf{Distance matrix} & \textbf{no norm.} & \textbf{norm.} \\
	\midrule
    Baseline & none (constant) & 68.47\% & 69.33\% \\ 
    Baseline & Euclidean & 68.94\% & 70.98\% \\ 
    Baseline & Euclidean$^2$ & 71.00\% & 71.64\% \\ 
    Cosine & cosine & 72.61\% & 73.36\% \\ 
    Cosine & cosine$^{1.6}$ & 73.12\% & 73.92\% \\
	\bottomrule
  \end{tabular}
 \vspace{0mm}
\end{table}

We have tried different lookup methods, such as direct comparison of subsequences or measuring edit distances. Out of the tested methods, the best results were obtained with dynamic time warping. We have also tried to post-process the quantizations, but all attempts worsened the results. This is probably due to loss of useful information, so we run DTW on vanilla quantizations.

For the sWUGGY test set, it was not possible to differentiate between \emph{base} and \emph{OOV}, so we have used only DTW.
For \emph{OOV} words, the difference between the results obtained with DTW and LSTM was minor. 

A significant improvement to our DTW relies on a technical detail. If we match the word correctly, we expect the match cost to be spread evenly over the entire sequence. However, when we match the word incorrectly, we expect the cost to be high in some places and low in the others. Thus, we increase the cost of distant pseudo-phones, and decrease the cost of similar ones by raising the cost in distance matrix to some power, which sharpens the distances. In our case, $1.6$ was the best for the cosine metric quantization and $2.0$ for the baseline quantization.

Processing of the train set took 18 hours on conventional hardware for both baseline and nullspace quantizations.
Results presented in Table \ref{tab:swuggy_dtw} show that both normalization and modification of the distance matrix yielded a significant improvement in the score.

\section{Semantic Task: sSIMI}
The goal of the task is to judge the semantic similarity of pairs of words (see \cite{baroni-etal-2014-dont}, \cite{schnabel2015evaluation}). That similarity is then compared 
with semantic judgments made by the human annotators, 
collected from a set of 13 existing semantic similarity and relatedness tests (including WordSimilarity-353\cite{FinkelsteinGMRSWR02a}, and mturk-771\cite{HalawiDGK12} which was used as a development set).

The submission format encouraged solutions which assign a vector at every temporal location, with a simple pooling method to aggregate them into a vector for the entire recording.
The pooling methods included \emph{min}, \emph{max}, \emph{avg}, \emph{last}, etc. In our submission, we have computed a vector for an entire recording, and replicated it along the time axis, so that after pooling with aforementioned functions it would remain unchanged.\\

\noindent\textbf{Preparation of the Corpus } We rely on the baseline pseudo-phone units, extracted with CPC and quantized with k-means. Streams of recognized pseudo-phones contain symbol repetitions, and we treat such blocks as higher order units. We further simplify these sequences by heuristically collapsing subsequent occurrences of the same pseudo-phone, and unify blocks which occur in similar contexts. The treat the result of this procedure as an approximation of a phoneme-level transcription with no word boundaries.\\

\noindent\textbf{Segmentation }
We apply segmentation into pseudo-words with SentencePiece~\cite{kudo_sentencepiece_2018}, which maximizes the probability under a unigram language model~\cite{kudo_subword_2018}. Given a vocabulary $\mathcal{V}={w_1,\ldots,w_n}$ with associated occurrence probabilities $(p_1,\ldots,p_n)$ and an utterance $x$, the most probable segmentation is determined with the Viterbi algorithm.
The vocabulary $\mathcal{V}$ is refined iteratively by maximizing the probability of every utterance under a unigram language model:
    $P(x) = \prod_i p(x_i)$.

We apply SentencePiece with target vocabulary size 50k.
By using ground-truth transcriptions, we found that the corpus has 18,705,420 words, which translates to 1.95 pseudo-word for every real word.\\


\noindent\textbf{Embedding and Retrieval }
With the corpus segmented into pseudo-words, we train an ordinary word2vec model~\cite{mikolov2013efficient}.
Every recording in the similarity task dataset comprises a single word, and we convert each of them to a sequence of pseudo-phones.
Some of those sequences exist in our pseudo-word vocabulary, and already have a unique word embedding calculated with word2vec. Others need to be built from smaller pseudo-word units. A simple way of doing so would be to split the sequence with SentencePiece into known pseudo-words. Knowing that the pseudo-representations tend to be noisy, we instead find the closest matches of each of them in the training corpus wrt. edit distance.
Then, word2vec embeddings of these matched pseudo-words are averaged to a single embedding for every input recording.

\subsection{Experiments for sSIMI}
Since our approach differs from the ZeroSpeech baseline one, we decided to present other word-oriented toplines for sSIMI, that better suit our approach. We compare word vectors trained on a large corpus, LibriSpeech transcriptions, and on our tokenization of LibriSpeech transcriptions. In that last case, we delete spaces and tokenize into 50k units using SentencePiece. The results are presented in Table~\ref{tab:simi_toplines}. The embeddings trained on word corpora outperform the RoBERTa topline, which suggests that the proposed approach might deserve further investigation.

\begin{table}[htb]
  \centering
  \caption{Toplines for word-based sSIMI (dev set). First three methods used all word pairs from the dev set, the result for RoBERTa\cite{liu2019roberta} are for the synthetized part of the data.}
  \label{tab:simi_toplines}
  \begin{tabular}{l c}  
    \toprule
  \textbf{Method}             & \textbf{synth.} \\   
  \midrule
  Google News word2vec                 & 65.5   \\ 
  LibriSpeech 960 transcriptions (w2v) & 36.3   \\ 
  Tokenized Librispeech                & 16.8   \\  
  RoBERTa (ZeroSpeech2021 Topline)     & 32.28  \\
  \bottomrule
 \end{tabular}
\end{table}

Table~\ref{tab:simi_results} presents our results in the ZeroSpeech contest together with other submissions. Our approach achieves the best score in the LibriSpeech subcategory, where the recordings are cut from LibriSpeech and not synthesized. This might indicate that our method is able to discover semantic shades of the known words from the corpus, but unable to generalize further.

\begin{table}[htb]
  \centering
  \caption{Correlation between human judgments and system responses ($\times$ 100). For other contestants the best submission on the test part of the data is presented.}
  \label{tab:simi_results}
  \begin{tabular}{l c c c c}  
    \toprule
    & \multicolumn{2}{c}{\textbf{synth.}} & \multicolumn{2}{c}{\textbf{libri.}} \\
  \textbf{Method}                      & \textbf{dev} &  \textbf{test} & \textbf{dev} & \textbf{test}  \\
  \midrule
  LSTM Baseline                        & 4.42       &  7.35      &  7.07       & 2.38   \\  
  BERT Baseline                        & 6.25       &  5.17      &  4.35       & 2.48   \\ 
  \midrule
  Ours                                 & 5.90       &  2.42      &  10.20      & 9.02   \\ 
  van Niekerk et al.                   & 4.29       &  9.23      &  7.69       & -1.14  \\ 
  Liu et al.                           & 3.16       &  7.30      & 1.79        & -4.33  \\ 
  Maekaku et al.                       & -2.10      &  6.74      & 8.89        & 2.03   \\ 
  \bottomrule
 \end{tabular}
\vspace{0mm}
\end{table}
\section{Syntactic Task: sBLIMP}

BLIMP~\cite{warstadt2020blimp} is a challenge dataset for evaluating to what extent language models can understand the English grammar. The dataset consists of pairs of similar sentences, and in every pair only one sentence is correct. In sBLIMP all sentences are synthesized. Since the aim of BLIMP was to evaluate how sentence likelihood is related to its grammatical correctness, there is a natural strategy of solving sBLIMP: use a language model to compute sentence probability and pick the most likely sentence from the pair as the correct one. To this end we use a LSTM language model trained on quantized nullspace features from LibriSpeech dev subset. In the competition, it had 53\% accuracy both on dev and test sets, slightly outperforming the baseline, and being close to 54\% of the best submission.

However, this result is not impressive at all: LSTM with random weights has 52.9\% accuracy. We hypothesize that this is caused by unbalanced utterance lengths in sBLIMP. We have discovered that incorrect sentences are typically longer than the correct ones.
Nevertheless, it is worth saying that BLIMP even in text version is definitely a non-trivial task, as for instance a big LSTM trained on large text corpus achieves only 70\% accuracy~\cite{gulordava-etal-2018-colorless}, and even large Transformer\cite{vaswani2017attention} model like GPT-2 don't exceed 82\% \cite{warstadt2020blimp, radford2019language}.

\section{Conclusions and Future Works}
We have presented a low-resource, information-retrieval based approach to the tasks of the Zero Resource Speech Challenge 2021. We were able to outperform baselines on every task, and achieve best or close to the best results on all four tasks. Still, many issues deserve further investigation.

First, we can explore the relationship between the ability of neural networks to memorize words, and contrast that with fuzzy information retrieval system. Is it possible to discover the dictionary from recordings, using some combinations of these approaches?

Moreover, we believe  that there is a potential synergy with computing semantic vector representation for pseudo-words using word2vec -- mainly because it is inexpensive to compute, and moving to bigger datasets can lead to a substantial improvement of embedding quality.

Solving BLIMP in the Zero Resource regime is undoubtedly an ambitious task. We believe that it is worth to consider its simpler, artificially created variants.
For instance, a variant in which the incorrect sentences were created by changing word order, or by replacing a randomly chosen word with another. Such simpler task can produce less noisy results.

Lastly, we can explore approach similar to averaging with centroids, but applied during training of CPC. For example by adding a loss based on the distance to simultaneously computed centroids, hoping that its denoising effect will improve the extracted representations.

\section{Acknowledgments}
The authors thank Polish National Science Center for funding
under the OPUS-18 2019/35/B/ST6/04379 grant and the PlGrid consortium for computational resources.

\bibliographystyle{IEEEtran}
\bibliography{mybib}

\begin{thebibliography}{10}
\providecommand{\url}[1]{#1}
\csname url@samestyle\endcsname
\providecommand{\newblock}{\relax}
\providecommand{\bibinfo}[2]{#2}
\providecommand{\BIBentrySTDinterwordspacing}{\spaceskip=0pt\relax}
\providecommand{\BIBentryALTinterwordstretchfactor}{4}
\providecommand{\BIBentryALTinterwordspacing}{\spaceskip=\fontdimen2\font plus
\BIBentryALTinterwordstretchfactor\fontdimen3\font minus
  \fontdimen4\font\relax}
\providecommand{\BIBforeignlanguage}[2]{{%
\expandafter\ifx\csname l@#1\endcsname\relax
\typeout{** WARNING: IEEEtran.bst: No hyphenation pattern has been}%
\typeout{** loaded for the language `#1'. Using the pattern for}%
\typeout{** the default language instead.}%
\else
\language=\csname l@#1\endcsname
\fi
#2}}
\providecommand{\BIBdecl}{\relax}
\BIBdecl

\bibitem{versteegh2015zero}
M.~Versteegh, R.~Thiolliere, T.~Schatz, X.~N. Cao, X.~Anguera, A.~Jansen, and
  E.~Dupoux, ``The zero resource speech challenge 2015,'' in
  \emph{Interspeech}, 2015.

\bibitem{nguyen_zero_2020}
T.~A. Nguyen, M.~de~Seyssel, P.~Rozé, M.~Rivière, E.~Kharitonov, A.~Baevski,
  E.~Dunbar, and E.~Dupoux, ``{The Zero Resource Speech Benchmark 2021: Metrics
  and baselines for unsupervised spoken language modeling},'' in
  \emph{Self-Supervised Learning for Speech and Audio Processing Workshop @
  NeurIPS}, 2020.

\bibitem{kahn2020libri}
J.~Kahn, M.~Rivi{\`e}re, W.~Zheng, E.~Kharitonov, Q.~Xu, P.-E. Mazar{\'e},
  J.~Karadayi, V.~Liptchinsky, R.~Collobert, C.~Fuegen \emph{et~al.},
  ``Libri-light: A benchmark for asr with limited or no supervision,'' in
  \emph{ICASSP 2020-2020 IEEE International Conference on Acoustics, Speech and
  Signal Processing (ICASSP)}.\hskip 1em plus 0.5em minus 0.4em\relax IEEE,
  2020, pp. 7669--7673.

\bibitem{7178964}
V.~{Panayotov}, G.~{Chen}, D.~{Povey}, and S.~{Khudanpur}, ``Librispeech: An
  asr corpus based on public domain audio books,'' in \emph{2015 IEEE
  International Conference on Acoustics, Speech and Signal Processing
  (ICASSP)}, 2015, pp. 5206--5210.

\bibitem{oord_representation_2018}
A.~van~den Oord, Y.~Li, and O.~Vinyals, ``Representation {{Learning}} with
  {{Contrastive Predictive Coding}},'' \emph{arXiv:1807.03748 [cs, stat]}, Jul.
  2018.

\bibitem{mikolov2013efficient}
T.~Mikolov, K.~Chen, G.~Corrado, and J.~Dean, ``Efficient estimation of word
  representations in vector space,'' 2013.

\bibitem{keuleers2010wuggy}
E.~Keuleers and M.~Brysbaert, ``Wuggy: A multilingual pseudoword generator,''
  \emph{Behavior research methods}, vol.~42, no.~3, pp. 627--633, 2010.

\bibitem{le-godais-etal-2017-comparing}
\BIBentryALTinterwordspacing
G.~Le~Godais, T.~Linzen, and E.~Dupoux, ``Comparing character-level neural
  language models using a lexical decision task,'' in \emph{Proceedings of the
  15th Conference of the {E}uropean Chapter of the Association for
  Computational Linguistics: Volume 2, Short Papers}.\hskip 1em plus 0.5em
  minus 0.4em\relax Valencia, Spain: Association for Computational Linguistics,
  Apr. 2017, pp. 125--130. [Online]. Available:
  \url{https://www.aclweb.org/anthology/E17-2020}
\BIBentrySTDinterwordspacing

\bibitem{1163055}
H.~{Sakoe} and S.~{Chiba}, ``Dynamic programming algorithm optimization for
  spoken word recognition,'' \emph{IEEE Transactions on Acoustics, Speech, and
  Signal Processing}, vol.~26, no.~1, pp. 43--49, 1978.

\bibitem{baroni-etal-2014-dont}
\BIBentryALTinterwordspacing
M.~Baroni, G.~Dinu, and G.~Kruszewski, ``Don{'}t count, predict! a systematic
  comparison of context-counting vs. context-predicting semantic vectors,'' in
  \emph{Proceedings of the 52nd Annual Meeting of the Association for
  Computational Linguistics (Volume 1: Long Papers)}.\hskip 1em plus 0.5em
  minus 0.4em\relax Baltimore, Maryland: Association for Computational
  Linguistics, Jun. 2014, pp. 238--247. [Online]. Available:
  \url{https://www.aclweb.org/anthology/P14-1023}
\BIBentrySTDinterwordspacing

\bibitem{schnabel2015evaluation}
T.~Schnabel, I.~Labutov, D.~Mimno, and T.~Joachims, ``Evaluation methods for
  unsupervised word embeddings,'' in \emph{Proceedings of the 2015 conference
  on empirical methods in natural language processing}, 2015, pp. 298--307.

\bibitem{FinkelsteinGMRSWR02a}
L.~Finkelstein, E.~Gabrilovich, Y.~Matias, E.~Rivlin, Z.~Solan, G.~Wolfman, and
  E.~Ruppin, ``Placing search in context: the concept revisited.'' \emph{ACM
  Trans. Inf. Syst.}, vol.~20, no.~1, pp. 116--131, 2002.

\bibitem{HalawiDGK12}
\BIBentryALTinterwordspacing
G.~Halawi, G.~Dror, E.~Gabrilovich, and Y.~Koren, ``Large-scale learning of
  word relatedness with constraints.'' in \emph{KDD}, Q.~Yang, D.~Agarwal, and
  J.~Pei, Eds.\hskip 1em plus 0.5em minus 0.4em\relax ACM, 2012, pp.
  1406--1414. [Online]. Available:
  \url{http://dblp.uni-trier.de/db/conf/kdd/kdd2012.html#HalawiDGK12}
\BIBentrySTDinterwordspacing

\bibitem{kudo_sentencepiece_2018}
\BIBentryALTinterwordspacing
T.~Kudo and J.~Richardson, ``{S}entence{P}iece: A simple and language
  independent subword tokenizer and detokenizer for neural text processing,''
  in \emph{Proceedings of the 2018 Conference on Empirical Methods in Natural
  Language Processing: System Demonstrations}.\hskip 1em plus 0.5em minus
  0.4em\relax Brussels, Belgium: Association for Computational Linguistics,
  Nov. 2018, pp. 66--71. [Online]. Available:
  \url{https://www.aclweb.org/anthology/D18-2012}
\BIBentrySTDinterwordspacing

\bibitem{kudo_subword_2018}
\BIBentryALTinterwordspacing
T.~Kudo, ``Subword regularization: Improving neural network translation models
  with multiple subword candidates,'' in \emph{Proceedings of the 56th Annual
  Meeting of the Association for Computational Linguistics (Volume 1: Long
  Papers)}.\hskip 1em plus 0.5em minus 0.4em\relax Melbourne, Australia:
  Association for Computational Linguistics, Jul. 2018, pp. 66--75. [Online].
  Available: \url{https://www.aclweb.org/anthology/P18-1007}
\BIBentrySTDinterwordspacing

\bibitem{liu2019roberta}
\BIBentryALTinterwordspacing
Y.~Liu, M.~Ott, N.~Goyal, J.~Du, M.~Joshi, D.~Chen, O.~Levy, M.~Lewis,
  L.~Zettlemoyer, and V.~Stoyanov, ``Roberta: A robustly optimized bert
  pretraining approach,'' 2019, cite arxiv:1907.11692. [Online]. Available:
  \url{http://arxiv.org/abs/1907.11692}
\BIBentrySTDinterwordspacing

\bibitem{warstadt2020blimp}
A.~Warstadt, A.~Parrish, H.~Liu, A.~Mohananey, W.~Peng, S.-F. Wang, and S.~R.
  Bowman, ``Blimp: The benchmark of linguistic minimal pairs for english,''
  \emph{Transactions of the Association for Computational Linguistics}, vol.~8,
  pp. 377--392, 2020.

\bibitem{gulordava-etal-2018-colorless}
\BIBentryALTinterwordspacing
K.~Gulordava, P.~Bojanowski, E.~Grave, T.~Linzen, and M.~Baroni, ``Colorless
  green recurrent networks dream hierarchically,'' in \emph{Proceedings of the
  2018 Conference of the North {A}merican Chapter of the Association for
  Computational Linguistics: Human Language Technologies, Volume 1 (Long
  Papers)}.\hskip 1em plus 0.5em minus 0.4em\relax New Orleans, Louisiana:
  Association for Computational Linguistics, Jun. 2018, pp. 1195--1205.
  [Online]. Available: \url{https://www.aclweb.org/anthology/N18-1108}
\BIBentrySTDinterwordspacing

\bibitem{vaswani2017attention}
A.~Vaswani, N.~Shazeer, N.~Parmar, J.~Uszkoreit, L.~Jones, A.~N. Gomez,
  L.~Kaiser, and I.~Polosukhin, ``Attention is all you need,'' 2017.

\bibitem{radford2019language}
A.~Radford, J.~Wu, R.~Child, D.~Luan, D.~Amodei, and I.~Sutskever, ``Language
  models are unsupervised multitask learners,'' 2019.

\end{thebibliography}

\end{document}